\documentclass[letterpaper, 10 pt, conference]{ieeeconf}  % Comment this line out if you need a4paper

\IEEEoverridecommandlockouts                              % This command is only needed if 
\usepackage{graphics} % for pdf, bitmapped graphics files
\usepackage{epsfig} % for postscript graphics files
\usepackage{mathptmx} % assumes new font selection scheme installed
\usepackage{times} % assumes new font selection scheme installed
\usepackage{amsmath} % assumes amsmath package installed
\usepackage{amssymb}  % assumes amsmath package installed

\usepackage{float}
\usepackage{orcidlink}
\usepackage{booktabs}
\usepackage{multirow}
\usepackage{caption}
\usepackage{graphicx} % 使用graphicx宏包
\usepackage{subcaption}
\usepackage[linesnumbered,ruled,vlined]{algorithm2e}
\usepackage{tabularx} % Add this in the preamble
\usepackage{booktabs}
\usepackage{multirow}
\usepackage{xcolor}
\usepackage{caption}
\usepackage{gensymb}
\usepackage{url}
\usepackage{epsfig} % for postscript graphics files
\usepackage{mathptmx} % assumes new font selection scheme installed
\usepackage{times} % assumes new font selection scheme installed
\usepackage{amsmath} % assumes amsmath package installed
\usepackage{amssymb}  % assumes amsmath package installed
\usepackage{array}
\usepackage{color, soul}
\usepackage{booktabs}
\usepackage{colortbl}
\usepackage{tcolorbox}
\usepackage{color,xcolor}
\usepackage{multirow}

\usepackage{float}
\usepackage{cite}
\usepackage{xcolor}
\usepackage{algpseudocode}
\usepackage[export]{adjustbox}
\usepackage{balance}
\usepackage{rotating}
\usepackage{graphicx}

\title{\LARGE \bf
PTQ4RIS: Post-Training Quantization for Referring Image Segmentation
}

\author{Xiaoyan Jiang$^{1*}$, Hang Yang$^{1*}$, Kaiying Zhu$^{2}$, Xihe Qiu$^{1\dagger}$, Shibo Zhao$^{3}$, Sifan Zhou$^{3\ddagger}$% <-this % stops a space
\thanks{* Equal contribution, $\ddagger$ Project Lead.}% <-this % stops a space
\thanks{$\dagger$ Corresponding author: {\tt\small qiuxihe1993@gmail.com}.}
\thanks{$^{1}$ School of Electronic and Electrical Engineering, Shanghai University of Engineering Science. $^{2}$ SenseTime. $^{3}$ Carnegie Mellon University.}
}

\begin{document}

\maketitle
\thispagestyle{empty}
\pagestyle{empty}

%%%%%%%%%%%%%%%%%%%%%%%%%%%%%%%%%%%%%%%%%%%%%%%%%%%%%%%%%%%%%%%%%%%%%%%%%%%%%%%%
\begin{abstract}

Referring Image Segmentation (RIS), aims to segment the object referred by a given sentence in an image by understanding both visual and linguistic information. However, existing RIS methods tend to explore top-performance models, disregarding considerations for practical applications on resources-limited edge devices. This oversight poses a significant challenge for on-device RIS inference. To this end, we propose an effective and efficient post-training quantization framework termed PTQ4RIS.
Specifically, we first conduct an in-depth analysis of the root causes of performance degradation in RIS model quantization and propose dual-region quantization (DRQ) and reorder-based outlier-retained quantization (RORQ) to address the quantization difficulties in visual and text encoders. Extensive experiments on three benchmarks with different bits settings (from 8 to 4 bits) demonstrates its superior performance. Importantly, we are the first PTQ method specifically designed for the RIS task, highlighting the feasibility of PTQ in RIS applications. Code and video are available at  \href{https://github.com/gugu511yy/PTQ4RIS}{https://github.com/gugu511yy/PTQ4RIS}.

\end{abstract}
%
% \begin{keywords}
% Model compression, referring image segmentation, post-training quantization
% \end{keywords}
%
 \section{INTRODUCTION}
Referring Image Segmentation (RIS) \cite{21} aims to predict a pixel-wise mask of the target object in an image guided by a corresponding text description, which has extensive applications in human-computer interaction via natural language \cite{51,52} and advanced driving systems \cite{30,53}. For example, in a domestic service robot, the integration of RIS capabilities allows the robot to swiftly and accurately segment objects based on verbal commands. This allows the robot to respond effectively to user inputs, such as identifying "the red cup on the kitchen counter". By enhancing the robot's ability to jointly understand natural language and images, RIS plays a crucial role in the development of intelligent robotic systems.

\begin{figure}[t]
    \centering
    \begin{subfigure}[b]{0.5\textwidth}
        \centering
        \includegraphics[width=0.9\textwidth]{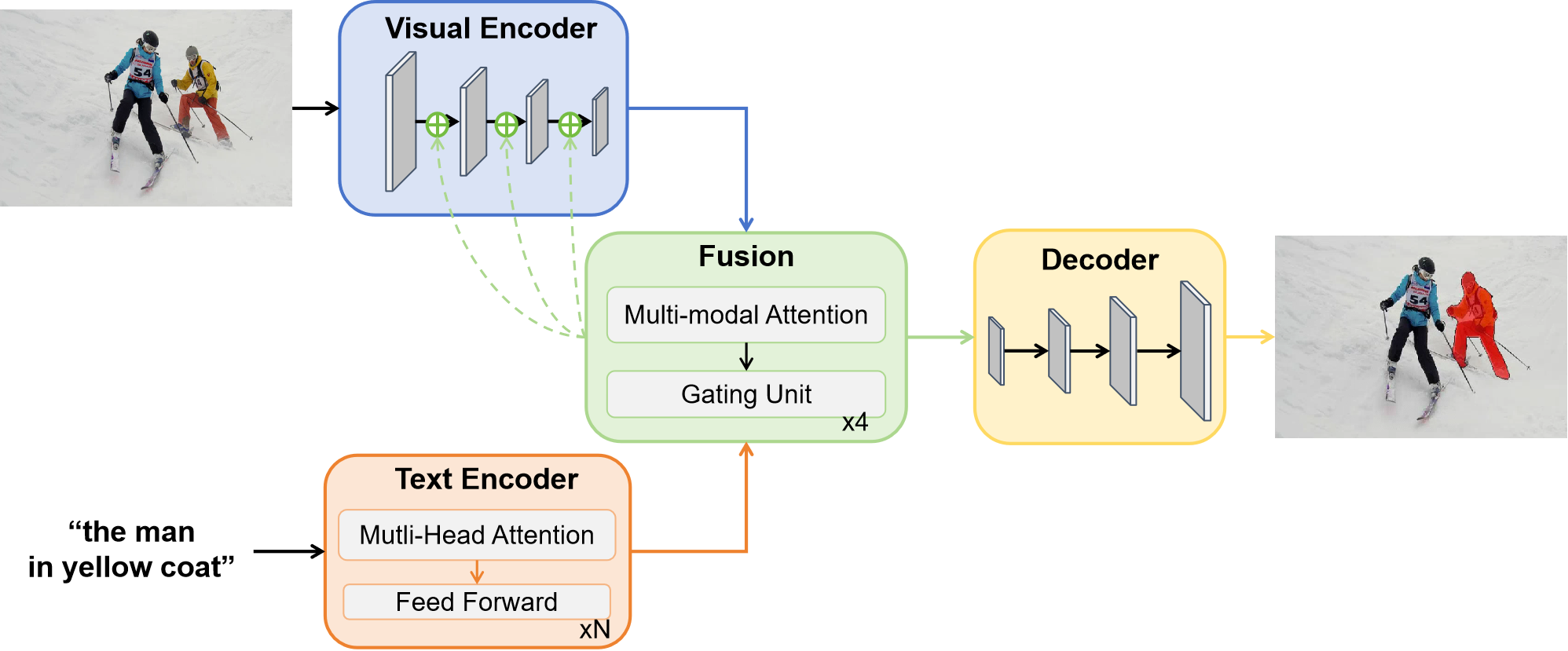} 
        \caption{}
        \label{fig:subfig1}
    \end{subfigure}    
    \begin{subfigure}[b]{0.5\textwidth}
        \centering
        \includegraphics[width=0.9\textwidth]{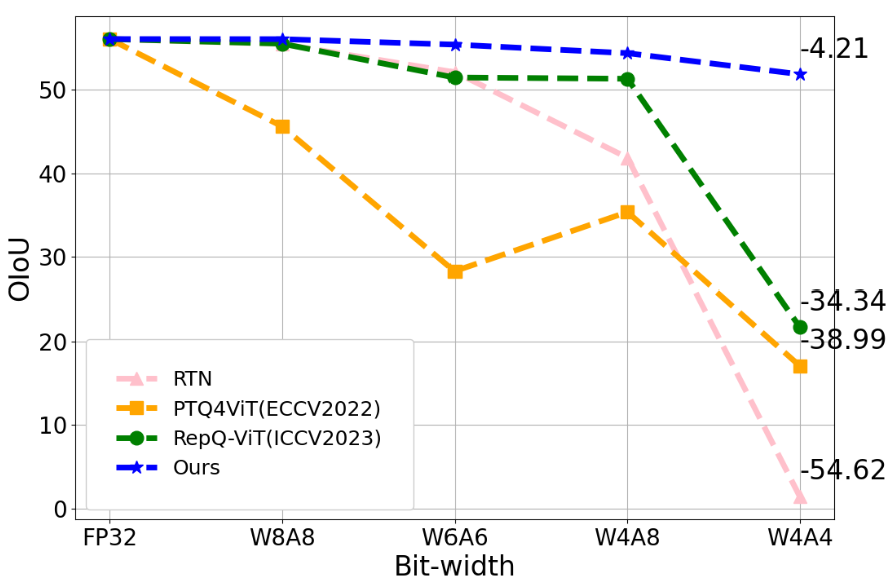} 
        \caption{}
        \label{fig:subfig2}
    \end{subfigure}
    
    \caption{(a) The pipeline of RIS, (b) Comparison of OIoU with various quantization methods (RTN, PTQ4ViT \cite{37}, RepQ-ViT \cite{38}) and our \textbf{PTQ4RIS} on the RefCOCO+ testB.}
    \label{fig:mainfig}
     \vspace{-8.0mm}
\end{figure}

As shown in Fig. \ref{fig:subfig1}, the existing RIS methods \cite{18,20,25} require modality-independent feature extraction and cross-modal feature fusion, ultimately producing dense pixel-level segmentation masks. Consequently, unlike conventional semantic or instance segmentation that rely solely on image data \cite{43,44}, the incorporation of natural language in RIS brings additional challenges in understanding the diverse vocabularies and syntactic varieties inherent in text modality. Naturally, the complex architecture design of RIS models inherently requires substantial computational resources and high-performance GPUs, significantly hindering their deployment on edge devices. For instance, the VLT model \cite{15} has 452M params and achieves only 50 ms inference latency on top-tier NVIDIA A100 GPUs. In contrast, such resource-intensive models pose significant challenges for deployment on the limited hardware resources onboard robot platforms.

Quantization is an efficient model compression approach \cite{45}, which converts weights and activation from 32-bit floating-point to 8-bit or lower integer fixed-point representation. This process significantly reduces model size and improves inference speed, thereby facilitating deployment on resource-constrained edge devices. In contrast to quantization-aware training (QAT) \cite{47,48} methods that demand access to a complete set of labeled training data and significant computational resources, post-training quantization (PTQ) \cite{32,33,36,lidar-ptq} is more advantageous for rapid and efficient industrial applications. PTQ requires only a small calibration set of unlabeled samples, eliminating the need to retrain the network using all available labeled data. This leads to a more streamlined and simple quantization process. Although several advanced PTQ methods have been proposed for RGB-based segmentation \cite{38,59} or language understanding tasks \cite{54,57}, applying it directly to RIS tasks inevitably leads to performance degradation due to the distribution difference between images and text.

In this paper, we aim to provide an efficient quantization solution for RIS task. As the first study in this area, we analyze each module of the RIS model individually and identify that the visual encoder and text encoder are particularly sensitive to quantization. For the visual encoder, we observe that the activation distribution after the Softmax and GeLU functions significantly deviates from a Gaussian distribution, leading to substantial quantization error when using a single scale. Therefore, we adopt a dual-region quantization \textbf{(DRQ)} method to quantize the activation in both ranges separately. For the text encoder, we find that the activation is significantly affected by numerous outliers, especially at lower bit widths. Thereby simply suppressing these outliers can result in a collapse of model performance. To this end, we propose a reorder-based outlier-retained quantization \textbf{(RORQ)} method, which iteratively partitions the activation values into groups and dynamically quantizes each group using distinct scale factors. For the quantization-friendly feature fusion and decoder modules, we directly apply a simple uniform quantization approach. 

Based on above findings and designs, we construct our \textbf{PTQ4RIS} framework, and evaluate it on three RIS benchmark datasets. 
Extensive experiments validate the effectiveness of our PTQ4RIS, as Fig. \ref{fig:subfig2} shows, highlighting its potential for deploying RIS models on onboard robot platforms. Notably, on the RefCOCO+ testB dataset, our PTQ4RIS achieves performance on par with the floating-point (FP) model under the W8A8 setting. Even in the W6A6 and W4A8 settings, the performance degradation is merely 0.66 OIoU and 1.54 OIoU, respectively. To our knowledge, this marks the first achievement of such impressive PTQ performance on RIS models. Our main contributions are summarized as follows:
\begin{itemize}
\item We unveil the root causes of performance collapse in the quantization of the RIS model, revealing that challenges primarily arise from the unique activation distributions of post-Softmax and post-GeLU in visual encoder, along with activation outliers in text encoder. 
\item We propose dual-region quantization (DRQ) and reorder-based outlier-retained quantization (RORQ) methods to tackle these issues. Furthermore, we introduce the PTQ4RIS framework, the first PTQ approach specifically designed for the RIS task.
\item Extensive experiments across various bit-width settings demonstrate the effectiveness and superiority of our PTQ4RIS. Specifically, the PTQ INT8 model's accuracy is almost the same as the FP32 model on partial datasets. 

\end{itemize}
\vspace{-1mm}
\section{RELATED WORK}
\noindent\textbf{Referring Image Segmentation.} 
The RIS task is first introduced in \cite{1}, using CNN \cite{63} and LSTM \cite{62} to extract visual and linguistic features, which are then connected for the final segmentation. Subsequent works \cite{2,4} enhance segmentation quality through recursive refinement networks (RRN) \cite{3} or dynamic networks. Others studies design strategies such as uni-directional \cite{6}, bi-directional fusion \cite{8,9}, and CGAN \cite{11} to capture mutual guidance between modalities. To address semantic complexity and ambiguity, some works \cite{5,7,10} integrate natural language processing knowledge to reason relationships between objects. With the success of Transformer \cite{29} in natural language processing and computer vision, recent studies \cite{20,21,22} employ encoder-decoder structures to enhance global contextual information. Some works \cite{15,16,17} align the visual and language features in the decoder after extracting them in the encoder. Notably, LAVT \cite{18} is a representative work that facilitates cross-modal interaction within the encoder, achieving impressive results. Additionally, some approaches \cite{19,23} address RIS as auto-regressive generation of polygon vertex coordinates, tackling complex shapes and occlusions. Leveraging parameter-efficient tuning, ETRIS \cite{26} introduces a cross-modal bridge through parameter-efficient fine-tuning, reducing backbone parameter while maintaining strong performance. Despite the advancement of RIS, their large parameters and computational burden still hinder effective deployment in robotic platforms. 

% \vspace{-2mm}
\noindent\textbf{Post-Training Quantization for CNN and Transformer.} 
Early post-training quantization (PTQ) methods such as AdaRound \cite{31}, BRECQ \cite{32}, QDrop \cite{33}, and PD-Quant \cite{34} are primarily developed for CNNs, employing techniques such as adaptive rounding, block-wise reconstruction, activation error correction, and distribution correction. However, these methods are not suitable for Transformers, prompting researchers to make significant efforts in developing effective quantization methods specifically for the transformer \cite{29} architecture. For Vision Transformer \cite{28}, PTQ-ViT \cite{35} adjusts bit-width using the nuclear norm of attention maps, ensuring feature mapping similarity before and after quantization. Based on this, FQ-ViT \cite{36} introduces a power-of-two factor and Log-Int-Softmax to achieve more comprehensive quantization. PTQ4ViT \cite{37} addresses the abnormal distributions of softmax and GeLU activation, while RepQ-ViT \cite{38} decouples quantization from inference to streamline operations. Additionally, there are quantization methods specifically designed for text-only transformers. For instance, I-BERT \cite{64} employs pure integer algorithms to quantize the entire inference process, while BiBERT \cite{56} introduces a fully binarized BERT model utilizing 1-bit weight and activation, significantly reducing computational costs and memory usage. 
Despite advancements in quantization methods for visual and textual modalities, the RIS task involves both images and text, rendering existing modality-specific quantization techniques inadequate. Furthermore, the cross-modal nature of RIS necessitates a tailored quantization approach to address the unique challenges of integrating visual and textual data within RIS models.

\section{PRELIMINARY}
\subsection{RIS Task Definition}
The input of the RIS task is a pair of an image \( I \in \mathbb{R}^{H \times W \times 3} \) and a natural language expression \( E \) that specifies an object in the image.
The model is ultimately tasked with generating a pixel-wise segmentation mask \( M \in \mathbb{R}^{H \times W} \) that delineates the object, assigning a binary label to each pixel in the image. The core challenge of this task is to enable the model to effectively process information from multiple modalities while accurately aligning linguistic and visual content, ensuring precise target localization and segmentation.

\subsection{Full-precision RIS Model} 
Due to the well-defined structure, compactness, and outstanding performance, the Language-Aware Visual Transformer (LAVT) \cite{18} model has garnered significant attention in RIS task. Consequently, we select representative LAVT as the full-precision baseline model for our quantization efforts. As Fig. \ref{fig:subfig1} shows, this network consists of four main components:

(1) \textbf{Visual Encoder:} For the input image \( I \), Swin Transformer \cite{66} is used to encode multi-scale visual feature maps \(V_i \in \mathbb{R}^{C_i \times H_i \times W_i}, i \in \{1,2,3,4\}\) from the corresponding stage. 
(2) \textbf{Text Encoder:} To extract linguistic features, a deep language expression model \cite{70} is employed to embed the input natural expression \(E\) into a high-dimensional word vector \( L\in \mathbb{R}^{C_t \times T}\). 
(3) \textbf{Fusion:} This stage includes two parts: the Pixel-Level Attention Mechanism (PWAM) and the Gating Unit. 
The \(V_i\) and \( L\) pass through the PWAM module, where a cross-modal attention mechanism fuses them to generate the multi-modal feature \(F_i \in \mathbb{R}^{C_i \times H_i \times W_i} \). The learnable gating unit then assigns weights to each element in \(F_i\).
%and combines it with \(V_i\), to produce a set of enhanced visual features embedded with linguistic information \(E_i \in \mathbb{R}^{C_i \times H_i \times W_i} \) , which are passed on to the next stage of Transformer layer for further processing.
%resulting in the visual feature \(E_i \in \mathbb{R}^{C_i \times H_i \times W_i} \) that enriches language information. 
(4) \textbf{Decoder:} \( F_i \) are progressively fused using two sets of 3x3 convolutions with ReLU activation, connected by BN layers, ultimately producing the segmentation mask \( M \).

\subsection{Weight and Activation Quantization}
Model quantization is one of the key technique for neural network compression. Given a floating-point (FP) vector $x$ (weights or activations), it can be uniformly quantized to $b$-bits as follows: 
% \vspace{-2.0mm}
\begin{align}
\label{eq:quant}
    \hat{\mathbf{x}} = \mathbf{Quant}_b(\mathbf{x},b) =(clamp(\lfloor \frac{\mathbf{x}}{s} \rceil+z, q_{min}, q_{max}) - z) \cdot s
\vspace{-3mm}
\end{align} 

where scale factor \( s \) is determined by the maximum and minimum values and the bit-width \( b\), reflecting the proportional relationship between FP values and integers. And $\text{clamp}()$ is used to clamp values in the range $(q_{\text{min}}, q_{\text{max}})$ and potentially causing clipping errors. \( z\) is the offset defined as the zero-point, in symmetric quantization, \( z\) is constrained to be 0. Here, wxcept visual encoder, we adopt uniform unsigned symmetric quantization, thereby $q_{min}=0$ and $q_{max}=2^{b} - 1$. In PTQ, weight quantization uses pre-trained model weights directly, while activation quantization employs a small calibration dataset to gather activation statistics, optimizing the quantization scheme to balance clipping and rounding errors effectively. 

% \vspace{-2.0mm}
\section{METHOD}
%\subsection{Overview}
In this section, we summarize the quantization challenges faced by the RIS model in Section \ref{sec:4.1}. The quantization method for the Visual Encoder is detailed in Section \ref{sec:4.2}, while the solution for activation outliers in the text encoder is presented in Section \ref{sec:4.3}. Finally, the overall algorithm of PTQ4RIS is described in Section \ref{sec:4.4}.

\subsection{Quantization Ristriction of RIS Model}
\label{sec:4.1}
Upon directly applying existing quantization methods to the FP RIS model, we observe a significant performance decline. Through in-depth analysis, we summarize the following limitations of RIS quantization:

\textbf{(1) Multi-modal hybrid CNN-transformer architectures:} 
Existing PTQ methods are developed for single-modal CNNs or transformers, which do not take into account the different parameter distributions inherent in hybrid architecture and cross-modal features. This mismatch leads to a substantial accuracy drop in RIS model. %when these methods are applied directly to

\textbf{(2) Non-normal distribution in post-Softmax and post-GeLU activations:} The performance degradation in the visual encoder is particularly pronounced, as the post-Softmax and post-GeLU activations exhibit distributions that are far from Gaussian. These values are crucial for guiding correlations between patches in the self-attention mechanism, making simple removal infeasible.

% \textbf{(3) Unignorable outliers in BERT linear layers:}
\textbf{(3) Unignorable outliers in text encoder:}
Since the semantic features extracted by the text encoder are critical inputs, any quantization errors in that section can accumulate and significantly impact the overall performance. Notably, the activation outliers in text encoder are particularly sensitive to quantization. These outliers vary irregularly with the input data and contain important information, direct culling will affect the semantic expression significantly.

To solve these challenges, we propose PTQ4RIS including dual-region quantization for vision encoder and reorder-based outlier-retained quantization for text encoder.

\vspace{-2mm}
\subsection{Dual-Region Quantization}
\label{sec:4.2}
Most PTQ quantization methods are based on Gaussian distributions \cite{37}. However, by visualizing the distributions of post-Softmax and post-GeLU activations, we find that the distributions are very special and cannot be quantized by uniform quantization methods \cite{34,46}. Firstly, the post-Softmax activations are between (0,1), most of the values are clustered near 0, and only a few large values approaching 1. In the self-attention, these values are mainly used for weighting patches.
The larger value means indicate a higher correlation between different patches, thus exerting a greater influence on the results.
Using a large scale factor \( s \) would reduce the quantization error of large values, whereas it would quantize a large portion of small values directly to 0. However, using a small scale factor \( s \) would significantly diminishes the correlation between patches.

\begin{figure}[h]
    \centering
    \vspace{2mm}
    \includegraphics[width=\linewidth]{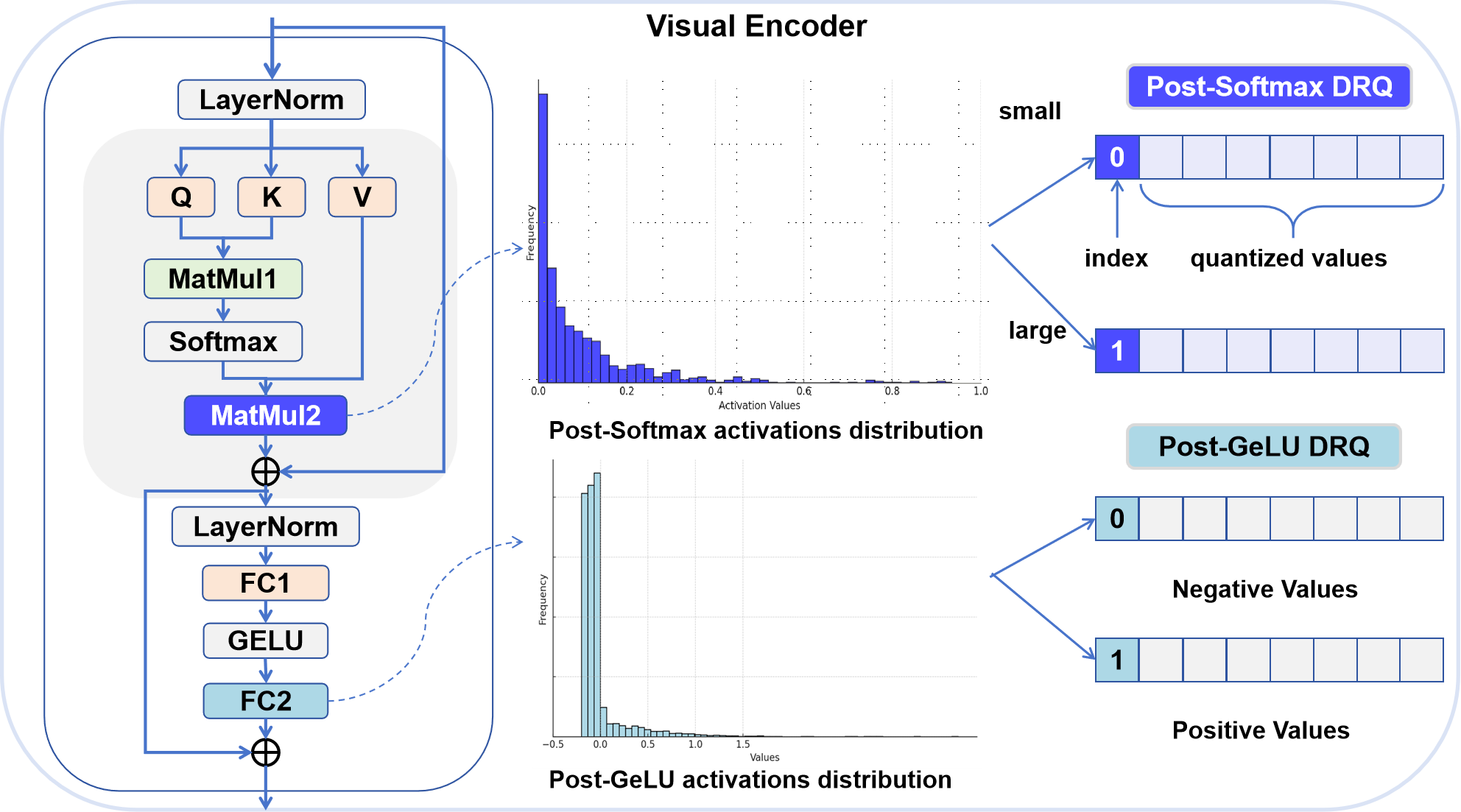}
    \caption{Dual-region quantization under 8-bit.}
    \label{fig:2}
    \vspace{-8.0mm}
\end{figure}

Moreover, the distribution of positive and negative intervals of activation values through the GELU function is highly asymmetric.
The positive distribution values are very large, but the negative values are highly close to 0. 
It is difficult to fit this distribution using symmetric quantization. 
Although non-uniform quantization can solve this problem, it is not applicable due to hardware \cite{65}.
Therefore, we propose \textbf{Dual-region Quantization} (DRQ) method, as shown in Fig. \ref{fig:2}. 
The function is described as follow:
\begin{equation}
\mathbf{T}_b(x, S_{R_1}, S_{R_2}) =
\begin{cases} 
\mathbf{Quant}_{b-1}(x, S_{R_1}), & x \in R_1 \\
\mathbf{Quant}_{b-1}(x, S_{R_2}), & \text{otherwise}~.
\end{cases}
\label{Eq:4}
\end{equation}

For post-Softmax values, we define two regions: \( R_1^s = [0, 2^{b - 1}S_{R_1^s}) \) and \( R_2^s = [0,2^{b - 1}S_{R_2^s}] \). To cover the whole range, we keep \( S_{R_2^s} = \frac{1}{2^b - 1} \) and large values can be well quantified in \(R_2^s\). 
For post-GeLU values, negative values in \( R_1^g = [-2^{b - 1}S_{R_1^g},0] \) and positive values in \( R_2^g = [0, 2^{b - 1}S_{R_2^g}] \). We also keep \( S_{R_1^g}\) fixed to make \( R_1^g \) just cover the entire range of negative numbers. When calibrating, we search for the best two scale factors, \( S_{R_1^s} \) and \( S_{R_2^g} \) to quantize activation values. Taking 8-bit quantization as an example, the final data format uses unsigned integers, including 1 bit for the region index and 7 bits for the quantized value. The region index is used for range calibration (0 for \(R_1\) , 1 for \(R_2\)). Meanwhile, we use \( S_{R_1} = 2^{-m} S_{R_2} \) to establish a constraint between the two regions, enabling the replacement of multiplication with left-shifting by $m$ bit. This approach facilitates rapid alignment of scale factors, thereby reducing the computational cost.
%Meanwhile, a shift operation is used to achieve fast alignment, where \( S_{R_1} = 2^{-m} S_{R_2} \).

The matrix multiplications \( QK^\top \)and \( PV \) in self-attention are critical operations in transformer,  where \( P = \text{softmax}((QK^\top)/\sqrt{d_k}) \) and  \( d_k \) is the dimension of the key vectors. For simplicity, we represent it as \( K = AB \). The Hessian-guided metric \cite{37} is used to determine the quantization scale factors \( S_A \) and \( S_B \) for the two matrices. Here, \( S_A \) represents the scale factor for matrices corresponding to either \( Q \) or \( P \).
We create the search space for \( S_A \) and \( S_B \) by linearly divided the ranges \([ \alpha \frac{A_{\text{max}}}{2^b - 1}, \beta \frac{A_{\text{max}}}{2^b - 1} ]\) and \([ \alpha \frac{B_{\text{max}}}{2^b - 1}, \beta \frac{B_{\text{max}}}{2^b - 1} ]\) to \( N \) candidates. We alternately search for the optimal scale factor in the above search space: First, by fixing \( S_B \), we determine the optimal \( S_A \) that minimizes the quantization error. Second, with \( S_A \) fixed, we search for the optimal \( S_B \). This process will repeated across several rounds within each layer.

\subsection{Reorder-based Outlier-Retained Quantization}
\label{sec:4.3}
\begin{figure}[h]
    \centering
    \includegraphics[width=\linewidth]{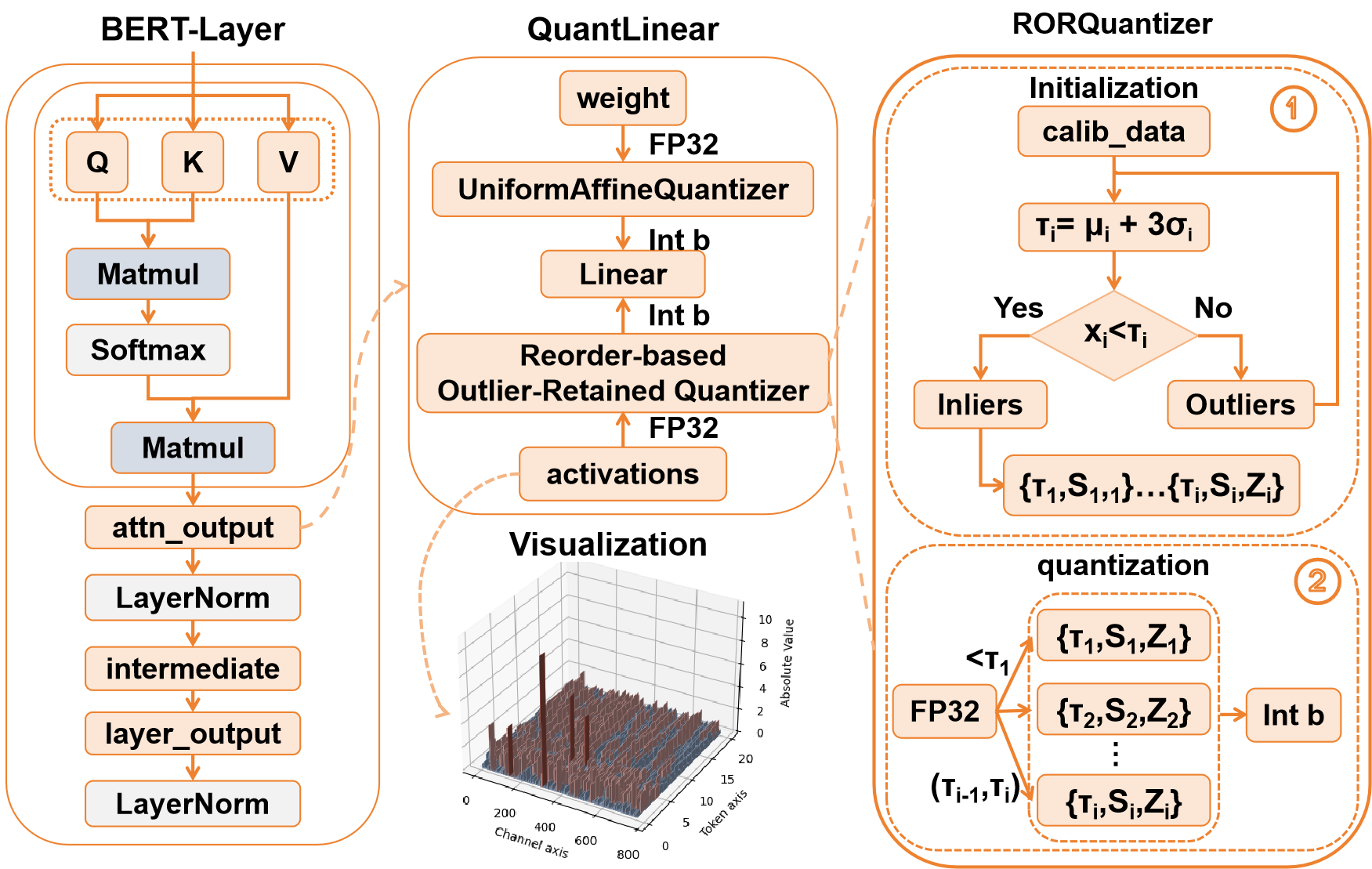}
    \caption{Reorder-based Outlier-Retained Quantization}
    \label{fig:4}
    \vspace{-4.0mm}
\end{figure}

In our in-depth analysis of the text encoding module, we identified significant outliers in the distribution of activations within the linear layers, as illustrated in Fig. \ref{fig:4}. Notably, these outliers cannot be directly excluded due to the presence of valid linguistic encoding information in the input segments. Capturing these activations with a single scale factor \( s \) results in severe quantization errors that accumulate and decrease the quantized model accuracy. Although per-channel quantization \cite{46} can alleviate this issue by assigning different scale factors to each channel, it requires specialized hardware support and adds a lot of computational overhead.

In order to strike a balance between efficiency and accuracy, we propose \textbf{Reorder-based Outlier-Retained Quantization} (RORQ), which enhances quantization precision by iteratively selecting outliers and calculating quantization parameters for different value groups. During the calibration phase, the RORQ determines the activation quantization parameters through the following steps:

\textbf{First}, the activations are converted to their absolute values to standardize the handling of both positive and negative values. Based on the empirical rule, we calculate the threshold \(\tau\) using Eq. \ref{eq:5}, which separates the outliers from the inliers:
% \begin{equation}
% \tau = \mu + 3\sigma~.
% \label{eq:5}
% \end{equation}
\begin{equation}
\tau = \mu + 3\sigma, \quad \mu = \frac{1}{N} \sum_{i=1}^{N} x_i, \quad \sigma = \sqrt{\frac{1}{N} \sum_{i=1}^{N} (x_i - \mu)^2}~. \label{eq:5}
\end{equation}
where \(\mu\) is the mean of the data distribution, and \(\sigma\) is the standard deviation, which measures the dispersion of data from the mean. Using \(3\sigma\) captures 99.7\% of the data, making the process of identifying outliers more robust and stable. Subsequently, we obtain the Inlier Set (data within the threshold) and the Outlier Set (data exceeding the threshold). \textbf{Second}, we employ the grid search technique of a uniform quantizer to determine the optimal quantization parameters for the current inlier set. \textbf{Third}, we recalculate the threshold of the current outlier set, dividing the data into new inlier and outlier groups, and determining the quantization parameters for the new inlier set. This process is repeated until the outlier set becomes empty or reaches a predefined maximum number of iterations. This iterative approach allows us to derive different quantization parameters for the activations of each group, effectively capturing the distribution of anomalies.

As a result, during the quantization process, the input data is dynamically grouped according to different thresholds and quantized using the corresponding quantization parameters of each group. Notably, our proposed approach, for unusually sensitive components, can flexibly adapt to the actual distribution of the input data, significantly reduce quantization errors, and thus improve quantized model performance.

\subsection{PTQ4RIS Algorithm}
\label{sec:4.4}
Except visual and text encoder, we also quantize other layers of the four modules of RIS model, such as convolutional layers and linear layers. Since they follow a Gaussian distribution, we use fine grain uniform quantization \cite{34,46} and search for the optimal scale factor for activations or weights layer by layer. The combined quantization of the four parts forms the PTQ4RIS framework in Algorithm \ref{ALog:1}.
\vspace{-3mm}
\begin{algorithm}[h]
\footnotesize
\caption{Pipeline of PTQ4RIS Framework}
\label{ALog:1}
\KwIn{Full-precision RIS model, calibration data}
\KwOut{Quantized RIS model with integer fixed-point}

\textbf{Step 1: Output $\mathbf{O}_l$ \& Gradient $\frac{\partial \mathcal{L}}{\partial \mathbf{O}_l}$ Collection}\\
\For{Each layer $l_i$ }{
    \If{$l_i$ in Visual Encoder}{
        Forward: compute $\mathbf{O}_l$\;
        Backward: compute $\frac{\partial \mathcal{L}}{\partial \mathbf{O}_l}$ for Hessian-guided metric\;
    }
}

\textbf{Step 2: Search Quantization Parameters $(s, z)$ }\\
Initialize quantized model with calibration data\;
\For{$l_i$ in Visual Encoder}{
    \If{Layer after Softmax or GeLU}{
        Define regions $R_1$ and $R_2$ for activations\;
        Search \( (S_{R_1^s}, S_{R_2^g}) \) using Hessian-guided metric\;
    }
    \ElseIf{Matmul}{
        Initialize search space for scale factors \( S_A \) and \( S_B \)\;
        \For{$r = 1$ to \#Rounds}{
            Fix \( S_B \), search for optimal \( S_A \)\;
            Fix \( S_A \), search for optimal \( S_B \)\;
        }
    }
}
\For{$l_i $ in Text Encoder}{
    \If{linear layers}{
        Apply RORQ \ref{sec:4.3} for activations to obtain $(\tau_i, s_i, z_i)$\;
    }
}
\For{$l_i $ in Fusion module and other text encoder layers}{
    Use Eq. \ref{eq:quant} for \( (s_{\text{weight}}, z_{\text{weight}}) \) and  \( (s_{\text{act}}, z_{\text{act}}) \)\;
}
\For{$l_i $ in Decoder}{
    Absorb BN layers into adjacent linear layers for efficient computation and compute \( (s_{\text{weight}}, z_{\text{weight}}) \) and  \( (s_{\text{act}}, z_{\text{act}}) \)\ based on Eq. \ref{eq:quant} with channel-wise setting;
}

\end{algorithm}

\vspace{-5.0mm}
\section{EXPERIMENTS}
\vspace{-1.0mm}
% \subsection{Dataset and Metrics}
\textbf{Datasets and Metrics:}
%Each dataset consists of three parts: original images, referring expressions, and pixel-level annotated binary ground truth maps. \textbf{Metrics:}
The experiments are conducted on three standard benchmark datasets: RefCOCO \cite{49}, RefCOCO+ \cite{49}, and G-Ref \cite{50}. Additionally, the G-Ref dataset has two different partitions, UMD and Google. Our experiments report on both partitions. Besides, we evaluate our proposed method using three metrics: Overall Intersection over Union (OIoU), Mean Intersection over Union (MIoU), and Precision\( @X,  X \in \{0.5, 0.7, 0.9\} \). 
\begin{table*}[ht]
\scriptsize
\caption{Quantization MIoU Results Across Different Datasets and Bit-widths (Full Precision is the ceiling performance)}
\label{1}
\centering
\renewcommand\arraystretch{1}
\resizebox{0.85\textwidth}{!}{%
    \begin{tabular}{c|c|c|c|c|c|c|c|c|c|c}
        \hline
        \raisebox{-2\height}{\textbf{Bit-width}} & \raisebox{-2\height}{\textbf{Method}} & \multicolumn{3}{c|}{\raisebox{-0.8\height}{\textbf{RefCOCO}}} & \multicolumn{3}{c|}{\raisebox{-0.8\height}{\textbf{RefCOCO+}}} & \multicolumn{3}{c}{\raisebox{-0.8\height}{\textbf{G-Ref}}} \\ 
        \cline{3-11} % Adds the horizontal line only for the columns 3-11
        & & val & testA & testB & val & testA & testB & val(U) & test(U) & val(G) \\
        \hline
        W32A32 & Full Precision & 74.31 & 76.63 & 70.61 & 66.69 & 71.47 & 60.01 & 65.91 & 66.01 & 64.08 \\
        \hline
        \multirow{4}{*}{W8A8} 
        & RTN & 73.28 & 76.06 & 69.46 & 65.70 & 70.43 & 58.95 & 65.44 & 65.60 & 63.86 \\
        & PTQ4ViT\cite{37} & 73.48 & 76.12 & 69.63 & 49.09 & 57.61 & 39.54 & 63.22 & 63.74 & 62.83 \\
        & RepQ-ViT\cite{38} & 73.51 & 75.98 & 69.81 & 66.32 & 71.20 & 59.62 & 65.18 & 65.36 & 63.50 \\
        \rowcolor{gray!20}
        & \textbf{PTQ4RIS (Ours)} & \textbf{73.54} & \textbf{76.24} & \textbf{70.21} & \textbf{66.42} & \textbf{71.32} & \textbf{59.76} & \textbf{65.47} & \textbf{65.62} & \textbf{63.93} \\
        \hline
        \multirow{4}{*}{W6A6} 
        & RTN & 69.47 & 74.44 & 68.16 & 62.33 & 66.99 & 54.76 & 62.88 & 63.29 & 60.91 \\
        & PTQ4ViT\cite{37} & 63.32 & 66.75 & 58.96 & 26.20 & 30.59 & 22.36 & 60.24 & 60.19 & 58.70 \\
        & RepQ-ViT\cite{38} & 65.16 & 68.25 & 63.71 & 58.67 & 65.27 & 50.81 & 61.15 & 61.55 & 58.76 \\
        \rowcolor{gray!20}
        & \textbf{PTQ4RIS (Ours)} & \textbf{72.85} & \textbf{75.37} & \textbf{68.62} & \textbf{65.10} & \textbf{69.70} & \textbf{58.55} & \textbf{65.02} & \textbf{65.31} & \textbf{63.66} \\
        \hline
        \multirow{4}{*}{W4A8} 
        & RTN & 62.80 & 65.30 & 58.98 & 44.01 & 47.54 & 39.26 & 53.71 & 53.89 & 41.31 \\
        & PTQ4ViT\cite{37} & 67.11 & 70.23 & 62.84 & 32.37 & 37.09 & 28.03 & 60.46 & 60.74 & 60.49 \\
        & RepQ-ViT\cite{38} & 67.82 & 71.09 & 63.50 & 59.00 & 65.36 & 51.08 & 62.19 & 62.74 & 60.50 \\
        \rowcolor{gray!20}
        & \textbf{PTQ4RIS (Ours)} & \textbf{72.62} & \textbf{75.12} & \textbf{68.47} & \textbf{64.29} & \textbf{69.50} & \textbf{57.45} & \textbf{63.87} & \textbf{63.96} & \textbf{62.08} \\
        \hline
        \multirow{4}{*}{W4A4} 
        & RTN & 5.88 & 5.89 & 4.34 & 0.86 & 0.96 & 0.92 & 2.09 & 2.17 & 0.53 \\
        & PTQ4ViT\cite{37} & 42.77 & 43.85 & 41.20 & 15.40 & 16.83 & 13.21 & 48.50 & 48.66 & 18.36 \\
        & RepQ-ViT\cite{38} & 28.89 & 29.51 & 40.10 & 22.55 & 25.86 & 18.00 & 54.68 & 54.84 & 48.04 \\
        \rowcolor{gray!20}
        & \textbf{PTQ4RIS (Ours)} & \textbf{69.53} & \textbf{71.67} & \textbf{65.35} & \textbf{61.25} & \textbf{65.77} & \textbf{54.29} & \textbf{60.60} & \textbf{60.86} & \textbf{59.92} \\
        \hline
    \end{tabular}%
}
\vspace{-3.0mm}
\end{table*}

% \vspace{-2.0mm}
% \subsection{Implementation Details}
\label{sec:5.2}

\textbf{Implementation Details:} Our full-precision model is LAVT\cite{18}. The calibration dataset number is set to 32, accounting for about \textbf{1\%} of the entire dataset. For the quantization design of the visual encoder, we set \(\alpha = 0.01\), \(\beta = 1.2\), \(N = 100\) to get search spaces, and the number of search round is set to 3. In the text endoer and fusion modules, the percentile methods \cite {69} is used to identify the scale factor of the matrix, otherwise use the mean square error (MSE) \cite {68} function, with channel-wise uniform quantization \cite{34} for weights and layer-wise uniform quantization for activations. For decoder, we use channel-wise uniform affine quantization for weights and activations. Our experiment video is available at \url{https://www.youtube.com/watch?v=EGy-PD7rRfk}.

\vspace{-2.0mm}
\subsection{Qualitative  Results}
Since there is no development PTQ method specifically designed for the RIS task, we implement a round-to-nearest quantization (RTN) to quantify our full-precision RIS model. Also, to compare with existing quantization methods, we re-implement two advanced well-known PTQ methods designed for vision transformer, namely PTQ4ViT \cite {37} and RepQ-ViT \cite {38}.
We conducte experiments using different weight-activation bit-widths for these methods. The results for MIoU and OIoU are shown in Table \ref{1} and Table \ref{2}.

When the bit width is 8, all four methods exhibit excellent performance, but our method shows superior results and performs nearly on par with the FP model on some datasets. 
For example, on RefCOCO testB dataset, we improve 0.75 MIoU and 0.36 OIoU compared to RTN, and 0.58 MIoU (0.29OIoU) and 0.40 MIoU (0.44OIoU) compared to PTQ4ViT and RepQ-ViT.
In W6A6 setting, the model performance after quantization using RTN is significantly better than the other two Transformer-based methods, indicating that the hybrid structure quantization method has the advantage at the lower bit width, but the least decrease in RTN is presented on the RefCOCO testA dataset at 2.19 MIoU and 2.28 OIoU, respectively. In contrast, our PTQ4RIS method maintains high performance, which improves 0.93 MIoU and 1.57 OIoU compared to RTN.

In W4A8 setting, taking RefCOCO+ val dataset as an example, the RTN performance decreases 22.68 MIoU (15.84 OIoU), while the PTQ4ViT is 34.32 MIoU (23.35 OIoU) and the RepQ-ViT is 7.69 MIoU (4.28 OIoU). Our method only decreases 2.40 MIoU (1.95 OIoU), which still fully demonstrates the advantages of the method. 
However, compared with the weight quantization set to 8 bits, the performance decrease of 2.13 MIoU (1.89 OIoU) is more obvious. This shows that the low weight bit width also affects the model performance, but compared with W4A4, it can be seen that the low bit quantization of the activation value is more sensitive. 
When the bit-width is W4A4, the other three methods almost lost performance, even though PTQ4ViT outperformed RTN and Rep-QViT, it has dropped by at least 17.41 MIoU (15.63 OIoU) on the val dataset of G-ref, but our PTQ4RIS only dropped by 5.31 MIoU and 3.67 OIoU, maintaining a high level of quantized model performance.

% %oiou
% % Table \ref{2} shows . In Table \ref{2}, .
\begin{table*}[h]
\scriptsize
\caption{Quantization OIoU Results Across Different Datasets and Bit-widths (Full Precision is the ceiling performance)}
\label{2}
\centering
\renewcommand\arraystretch{1}
\resizebox{0.85\textwidth}{!}{%
    \begin{tabular}{c|c|c|c|c|c|c|c|c|c|c}
        \hline
        \raisebox{-2\height}{\textbf{Bit-width}} & \raisebox{-2\height}{\textbf{Method}} & \multicolumn{3}{c|}{\raisebox{-0.8\height}{\textbf{RefCOCO}}} & \multicolumn{3}{c|}{\raisebox{-0.8\height}{\textbf{RefCOCO+}}} & \multicolumn{3}{c}{\raisebox{-0.8\height}{\textbf{G-Ref}}} \\ 
        \cline{3-11} % Adds the horizontal line only for the columns 3-11
        & & val & testA & testB & val & testA & testB & val(U) & test(U) & val(G) \\
        \hline
        W32A32 & Full Precision &  72.72 & 75.74 & 68.56 & 63.38 & 68.73 & 56.08 & 62.65 & 64.10 & 60.85 \\
        \hline
        \multirow{4}{*}{W8A8} 
        & RTN & 72.08 & 75.33 & 67.90 & 62.84 & 67.99 & 55.49 & 62.38 & 63.80 & 60.56 \\
        & PTQ4ViT\cite{37} & 72.07 & 75.33 & 67.97 & 53.85 & 60.99 & 45.64 & 61.06 & 62.78 & 60.20 \\
        & RepQ-ViT\cite{38} & 72.26 & 75.14 & 67.82 & 63.09 & 68.58 & 55.53 & 62.09 & 63.47 & 60.49 \\
        \rowcolor{gray!20}
        &\textbf{PTQ4RIS (Ours)} & \textbf{72.33} & \textbf{75.38} & \textbf{68.26} & \textbf{63.32} & \textbf{68.72} & \textbf{56.07} & \textbf{62.46} & \textbf{63.84} & \textbf{60.80} \\
        \hline
        \multirow{4}{*}{W6A6} 
        & RTN & 68.10 & 73.46 & 66.06 & 60.23 & 65.02 & 52.12 & 60.16 & 61.53 & 57.92 \\
        & PTQ4ViT\cite{37} & 65.03 & 68.64 & 59.97 & 31.90 & 35.91 & 28.32 & 57.75 & 58.71 & 57.74 \\
        & RepQ-ViT\cite{38} & 67.37 & 70.56 & 64.59 & 58.97 & 64.91 & 51.48 & 59.50 & 60.51 & 56.34 \\
        \rowcolor{gray!20}
        & \textbf{PTQ4RIS (Ours)} & \textbf{71.90} & \textbf{75.03} & \textbf{67.40} & \textbf{62.50} & \textbf{67.52} & \textbf{55.42} & \textbf{62.01} & \textbf{63.47} & \textbf{60.61} \\
        \hline
        \multirow{4}{*}{W4A8} 
        & RTN & 64.73 & 67.70 & 60.47 & 47.54 & 51.57 & 41.88 & 54.82 & 56.21 & 44.84 \\
        & PTQ4ViT\cite{37} & 68.18 & 71.98 & 63.58 & 40.03 & 44.32 & 35.38 & 58.94 & 59.92 & 58.32 \\
        & RepQ-ViT\cite{38} & 68.59 & 71.88 & 64.22 & 59.10 & 65.12 & 51.33 & 60.00 & 61.35 & 58.40 \\
        \rowcolor{gray!20}
        & \textbf{PTQ4RIS (Ours)} & \textbf{71.48} & \textbf{74.60} & \textbf{66.53} & \textbf{61.43} & \textbf{67.17} & \textbf{54.39} & \textbf{61.29} & \textbf{62.38} & \textbf{59.68} \\
        \hline
        \multirow{4}{*}{W4A4} 
        & RTN & 8.68 & 8.53 & 6.56 & 1.49 & 1.66 & 1.46 & 2.95 & 3.24 & 0.61 \\
        & PTQ4ViT\cite{37} & 47.03 & 48.54 & 45.36 & 15.40 & 19.42 & 17.09 & 47.02 & 48.31 & 21.00 \\
        & RepQ-ViT\cite{38} & 32.73 & 33.23 & 43.54 & 25.52 & 26.97 & 21.74 & 52.21 & 53.02 & 47.17 \\
        \rowcolor{gray!20}
        & \textbf{PTQ4RIS (Ours)} & \textbf{69.21} & \textbf{72.00} & \textbf{64.11} & \textbf{59.44} & \textbf{64.26} & \textbf{51.87} & \textbf{58.98} & \textbf{60.37} & \textbf{58.27} \\
        \hline
    \end{tabular}%
}
\vspace{-5.0mm}
\end{table*}

\vspace{-2.0mm}
\subsection{Ablation Studies}
We perform ablation experiments on the RefCOCO validation set with W4A4 setting.
\vspace{-2.0mm}
\begin{table}[H]
\centering
\caption{Ablation results of different quantized modules.}
\label{3}
\resizebox{\columnwidth}{!}{ % Adjusts the table to fit the column width
\begin{tabular}{c|c|c|c|c|c|c}
\toprule
\multirow{2}{*}{\textbf{Bits}} & \textbf{Visual} & \textbf{Text} & \textbf{Fusion} & \textbf{Decoder} & \multirow{2}{*}{\textbf{MIoU}} & \multirow{2}{*}{\textbf{OIoU}} \\
 & \textbf{DRQ} & \textbf{RORQ} & \textbf{RTN} & \textbf{RTN} &  &  \\
\hline
\multirow{6}{*}{W4A4} & - & - & - & - & 5.88 & 8.68 \\
& - & - & \checkmark & \checkmark & 19.77 & 21.15 \\
& \checkmark & - & - & - & 53.30 & 52.99 \\
& - & \checkmark & - & - & 21.21 & 24.28 \\
& \checkmark & \checkmark & - & - & 68.87 & 68.84 \\
\rowcolor{gray!20}
& \checkmark & \checkmark & \checkmark & \checkmark & \textbf{69.53} & \textbf{69.21} \\
\bottomrule
\end{tabular}
}
\vspace{-3.0mm}
\end{table}
\noindent\textbf{Ablation of different quantized modules.} Table \ref{3} shows evaluation of the effectiveness of the four modules (Visual Encoder, Text Encoder, Fusion, and Decoder) in our PTQ4RIS. It can be observed that the performance of the quantized model is very poor when using the naive RTN method. The use of a more fine grain search quantization strategy in the Fusion and Decoder sections results in an improvement of 13.89 MIoU and 12.47 OIoU. After using our DRQ method in Visual Encoder, MIoU and OIoU are improved from 5.88 and 8.68 to 53.30 and 52.99. Using RORQ in text encoder improves model performance by 15.33 MIoU (15.6 OIoU). 
It fully demonstrates the superior performance of each of our proposed method components. 

% re-order
\begin{table}[h]
\vspace{-2.0mm}
\centering
\scriptsize
\caption{Ablations of text encoder quantization.}
\label{4}
\resizebox{\columnwidth}{!}{ % Adjusts the table to fit the column width
\begin{tabular}{l|l|l|l|l|l}
\toprule
\textbf{Method} & \textbf{P@0.5} & \textbf{P@0.7} & \textbf{P@0.9} & \textbf{OIoU} & \textbf{MIoU} \\
\hline
Full Prec. & 84.63 & 75.25 & 34.56 & 72.72 & 74.31 \\
Per-tensor & 51.05 & 40.35 & 13.73 & 49.68 & 46.18 \\
Per-channel & 82.19 & 72.96 & 33.04 & 70.00 & 72.27 \\
Mean Division & 82.83 & 72.79 & 32.52 & 71.19 & 72.62 \\
Median + MAD & 77.93 & 68.15 & 29.59 & 67.11 & 68.97 \\
Confidence & 79.00 & 68.75 & 29.43 & 69.30 & 69.32 \\
\rowcolor{gray!20}
\textbf{Mean + 3SD} & \textbf{83.22} & \textbf{73.80} & \textbf{33.33} & \textbf{72.00} & \textbf{73.13} \\
\bottomrule
\end{tabular}
}
\vspace{-4.0mm}
\end{table}

\noindent\textbf{Ablation of RORQ.} Table \ref {4} shows the results of using different quantization methods in \textbf{Reorder-based Outlier-Retained Quantization} (RORQ). We compare per-tensor and per-channel quantization, as well as the RORQ using different computational threshold methods, including Mean Division, Median + Median Absolute Deviation (MAD), confidence level, and Mean + Three Standard Deviations (3SD). The results show that our RORQ method is better than coarse-grained quantization, and the threshold calculation method we used has the best performance.

\vspace{-2mm}
\section{CONCLUSION}
\vspace{-1mm}
In this paper, we propose a novel PTQ scheme specifically designed for the RIS task, called \textbf{PTQ4RIS}, aiming at enabling the deployment of large multi-modal models on edge devices. We analyze the data distribution and network structure of various modules and develop tailored quantization strategies for different components. Specifically, we propose the \textbf{Dual-Region Quantization} (DRO) for visual encoder and the \textbf{Reorder-based Outlier-Retained Quantization} (RORQ) for text encoder. The proposed PTQ4RIS method achieve superior performance under 8-bit quantization, and even under 6-bit and 4-bit settings, where other advanced methods experience significant performance degradation. We hope our approach can advance the development of RIS tasks on robotics. In the future, we will explore lower-bit quantization methods to further investigate the practical application potential of RIS model on robotic systems.

\bibliographystyle{IEEEtran}
\bibliography{IEEEabrv,strings,refs}

\end{document}